\documentclass[journal]{IEEEtran}

\usepackage{amsmath,amsfonts,amssymb}
\usepackage{algorithm}
\usepackage{algpseudocode}
\usepackage{array}
\usepackage{multirow}
\usepackage{booktabs}
\usepackage[caption=false,font=normalsize,labelfont=sf,textfont=sf]{subfig}
\usepackage[table]{xcolor}
\usepackage{textcomp}
\usepackage{stfloats}
\usepackage{url}
\usepackage{verbatim}
\usepackage{graphicx}
\usepackage{cite}
\usepackage{hyperref}

\begin{document}

\title{Omni-IGN: Onboard Image-Goal Navigation for Drones with Omnidirectional Vision}

\author{Zichen Yan, Rui Huang, Shenao Wang, Yucong Wang, Nuthasith Gerdpratoom, Lin Zhao\textsuperscript{*}
\thanks{The authors are with the Department of Electrical and Computer Engineering, National University of Singapore, Singapore.
         Email: zichenyan@u.nus.edu%
}
 \thanks{\textsuperscript{*}Corresponding author: Lin Zhao (E-mail: zhaolin@nus.edu.sg).}
 }

\maketitle

\begin{abstract}
Image-goal navigation (ImageNav) requires an agent to reach a location specified by a goal image in an unknown environment. For drones, this task demands exploration, goal-directed navigation, and collision avoidance under limited onboard computing resources. Existing aerial ImageNav methods are built on single-view observations, requiring additional rotation and exploration to search for goal-relevant cues outside the field of view. In this paper, we present Omni-IGN, a fully onboard ImageNav system that integrates quad-fisheye perception, an attention-based reinforcement learning (RL) policy, and a collision-aware safety layer. The policy adaptively fuses goal-relevant features across multiple views to provide directional guidance for continuous navigation. The quad-fisheye RGB images support both ImageNav and obstacle avoidance without a dedicated depth sensor. On Habitat benchmarks, Omni-IGN achieves the best performance among continuous-action baselines. Moreover, flight tests in IsaacSim validate the effectiveness of integrating learning-based depth estimation with the action safety layer for collision avoidance. Finally, real-world experiments demonstrate zero-shot sim-to-real transfer without any fine-tuning or external localization. The code and model checkpoints will be released upon acceptance.
\end{abstract}


\section{Introduction}
\IEEEPARstart{A}{erial} ImageNav requires a drone to search an unknown 3D environment and reach the location where a goal image was captured. Specifying destinations through images enhances drone autonomy and provides greater flexibility in mission execution, as the drone can discover a route without requiring an operator to map the environment or prescribe a trajectory in advance. This capability can reduce manual preparation in unknown or unstructured environments, enabling rapid deployment for time-critical tasks such as disaster rescue.

In this work, we investigate fully onboard omnidirectional ImageNav for drones, aiming to exploit broad visual coverage for more efficient goal search. Observations from all directions help capture goal-relevant cues beyond the forward field of view (FoV), reducing unnecessary rotations and repeated exploration~\cite{an2024etpnav, PanoNav}. Yet the central challenge is to turn this panoramic view into timely, goal-directed flight commands. Goal-relevant cues may occupy only one view, remain visually ambiguous, or be absent during exploration. Processing all views indiscriminately increases both computational cost and visual distraction. An effective policy must therefore extract useful goal-relevant cues while retaining the spatial context needed for exploration, all within limited onboard computational resources.

Existing studies address only parts of this problem under more limited task settings. Prior aerial ImageNav methods rely on single-view observations~\cite{gao2026flying,yan2025sign,zhu2026uanwm}. For drones equipped with omnidirectional vision, most studies instead target individual capabilities, including mapping and localization~\cite{liu2024omninxt}, exploration~\cite{wang2026palm}, and obstacle avoidance~\cite{zhang2026fly360}. OMNI-DRL~\cite{artizzu2022omni} combines omnidirectional perception with point-goal navigation and obstacle avoidance. However, ImageNav provides no compass guidance toward the target. The drone must actively explore the environment to discover the goal while avoiding collisions. Exploration, goal seeking, and collision avoidance are therefore tightly coupled. Real-world deployment further requires this coordination to remain robust to sim-to-real visual discrepancies that may affect both goal recognition and obstacle perception.

To enable efficient inference and smooth control, an end-to-end pipeline that directly maps visual observations to continuous velocity commands is necessary. Compared with the discrete action space used in TSGM~\cite{tsgm}, continuous control enables smoother motion and more responsive maneuvering, making it more compatible with drone dynamics. SIGN~\cite{yan2025sign} demonstrates velocity control for aerial ImageNav and highlights the increased training difficulty introduced by continuous actions, but considers only a single camera view and runs in an offboard mode. Our setting involves both high-dimensional panoramic observations and continuous control, increasing the difficulty of extracting goal-relevant visual representations while simultaneously learning effective exploration behaviors.

Collision avoidance presents an additional challenge for real-world deployment. Existing ImageNav benchmarks prioritize goal-reaching performance but neglect collisions. Although minor collisions may be tolerable for ground robots, even brief contact with obstacles can destabilize a drone and lead to crashes. Depth provides a direct measurement for assessing collision risk, but obtaining it through a dedicated depth sensor adds weight, cost, and hardware complexity. We therefore seek to integrate depth-based collision avoidance into aerial ImageNav while keeping the sensing configuration compact and lightweight. Achieving the aforementioned capabilities within limited onboard resources presents a further system-level challenge.

\begin{figure*}[!t]
    \centering
    \includegraphics[width=0.99\linewidth]{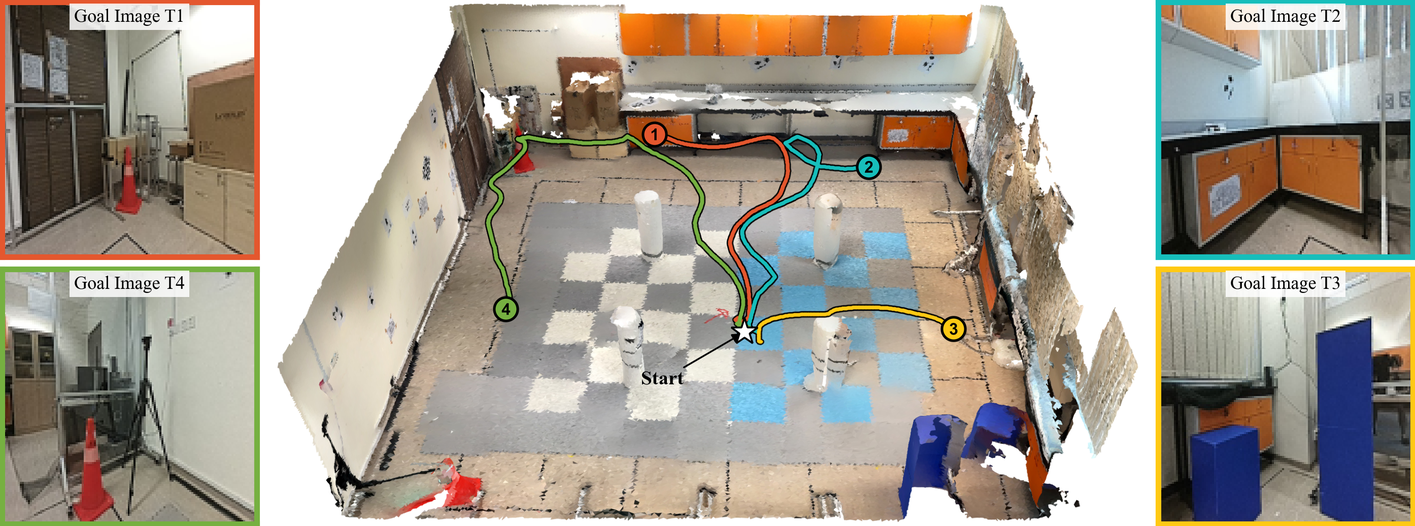}\\[2pt]
    \includegraphics[width=1.0\linewidth]{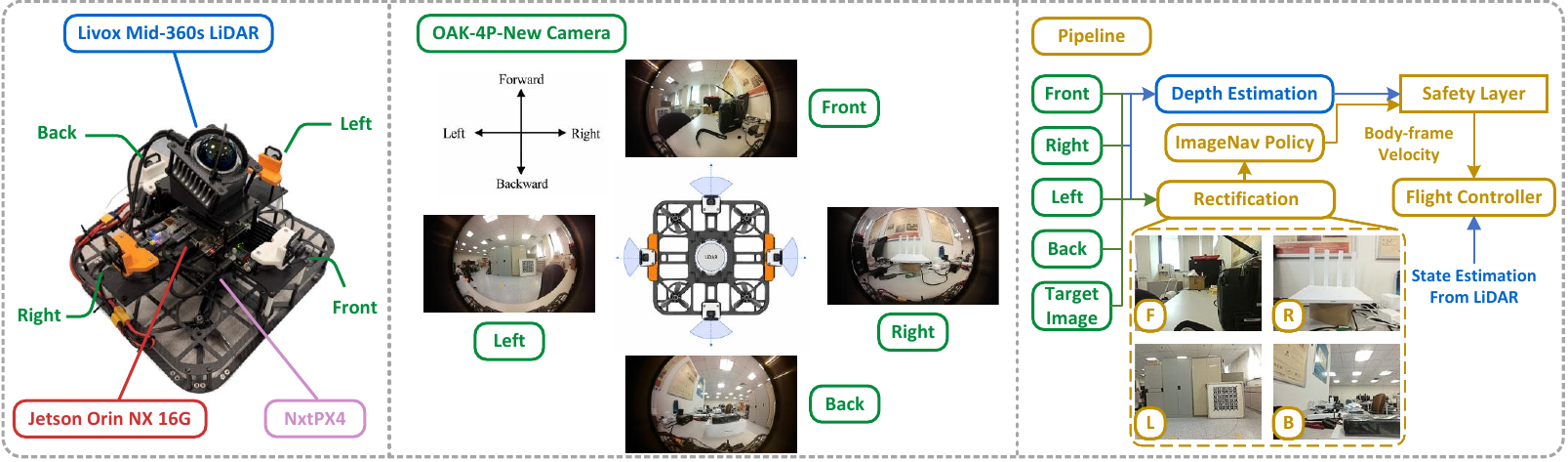}
    \caption{System design and data flow of Omni-IGN. The drone is equipped with a Jetson Orin NX 16G onboard computer, an OAK-4P-New quad-fisheye camera rig, and a Livox Mid-360s LiDAR. Rectified images from the four fisheye cameras are fed to the RL navigation policy, while overlapping views between the front camera and its left and right neighbors are used for stereo depth estimation and obstacle avoidance. The velocity command passes through a safety layer before being sent to the flight controller. Meanwhile, LiDAR provides state estimates required for closed-loop control.}
    \label{fig:title}
\end{figure*}

To address the above challenges, we present \textbf{Omni-IGN} (\textbf{Omni}directional \textbf{I}mage-\textbf{G}oal \textbf{N}avigation), a fully onboard aerial ImageNav system for which we design a shared quad-fisheye RGB sensing rig and integrate an end-to-end RL policy with a collision-aware safety layer supported by stereo depth estimation (see Fig.~\ref{fig:title}). We jointly encode each of the four camera views with the goal image and introduce a self-attention module (see Fig.~\ref{fig:model}) to adaptively fuse the resulting features for continuous velocity prediction. In parallel, we estimate stereo depth from overlapping views of our camera rig and use a collision-aware safety layer to correct unsafe actions before sending them to the flight controller. This design couples goal-conditioned navigation with geometric action correction while keeping both functions within a fully onboard pipeline.

Our main contributions are as follows:
\begin{itemize}
    \item To the best of our knowledge, Omni-IGN is the first fully onboard implementation of omnidirectional ImageNav on drones. We introduce a self-attention module that adaptively weights goal-conditioned visual features across four views, providing directional guidance for the RL policy and significantly reducing repeated exploration.

    \item We employ a quad-fisheye RGB rig shared by ImageNav and collision avoidance, and integrate stereo depth estimation from overlapping views with a safety layer that corrects unsafe actions. This safety mechanism integrates seamlessly with the continuous-action ImageNav policy without requiring a dedicated depth sensor.

    \item We evaluate Omni-IGN through simulation benchmarks and real-world flight experiments. Omni-IGN achieves state-of-the-art (SOTA) performance among continuous-action baselines on Habitat benchmarks~\cite{gibson,mp3d,hm3d}. IsaacSim validates the reliability of depth estimation and safety layer for collision avoidance, while real-world flights demonstrate successful zero-shot sim-to-real transfer without fine-tuning.
\end{itemize}

\section{Related Works}
\subsection{End-to-end ImageNav Algorithms}
Existing ImageNav algorithms use either single-view or panoramic observations. 
Single-view methods adopt diverse learning paradigms, including reinforcement learning (RL)~\cite{zer,ovrlv2,fgp}, goal-conditioned supervised learning~\cite{sridhar2024nomad, ViNT}, and world-model approaches~\cite{bar2025navigation}. 
To reduce unnecessary exploration and improve path efficiency, VGM~\cite{vgm} and TSGM~\cite{tsgm} use panoramic observations to build graph-based memories of visual and semantic information, respectively. Mem-Aug~\cite{memaug} instead uses episodic memory and attention to retrieve information from previously visited states. However, these panoramic methods adopt discrete actions that hinder smooth and continuous control. Processing multiple views also increases the computational burden, making it challenging to meet real-time requirements during onboard inference.

\subsection{Aerial ImageNav} 
Beyond conventional benchmark metrics, Aerial ImageNav imposes additional requirements for continuous control and real-time execution. Gao \textit{et al.}~\cite{gao2026flying} extend InstanceImageNav to drones using a hierarchical framework that builds cross-graph semantic memory, selects informative viewpoints for exploration, and generates dynamically feasible trajectories for navigation. SIGN~\cite{yan2025sign} develops an end-to-end RL policy that predicts velocity commands and integrates depth-based safety correction for real-world drone navigation. Zhu \textit{et al.}~\cite{zhu2026uanwm} propose an uncertainty-aware world model for outdoor aerial ImageNav. It evaluates candidate trajectories generated by a planner to enable robust navigation under model uncertainty. 

Despite their different paradigms, these methods primarily rely on front-view visual observations. In contrast, Omni-IGN introduces omnidirectional perception for aerial ImageNav, providing broader visual coverage while retaining continuous control suitable for onboard deployment.

\subsection{Drone Navigation with Omnidirectional Perception}
To overcome the limited FoV of conventional cameras, omnidirectional perception is explored for various drone applications. OmniNxt~\cite{liu2024omninxt} employs a quad-fisheye camera for mapping and localization. Wang \textit{et al.}~\cite{wang2026palm} combine quad-fisheye vision with a sparse topological map to guide autonomous exploration. Fly360~\cite{zhang2026fly360} further exploits panoramic perception for omnidirectional obstacle avoidance and local motion control.
OMNI-DRL~\cite{artizzu2022omni} studies RL-based point-goal navigation on drones with omnidirectional images, demonstrating improved navigation performance over limited-FoV observations in simulated forests. Its policy receives relative goal-direction information and selects discrete steering actions. In contrast, Omni-IGN supports exploration and ImageNav without access to oracle goal information, and generates continuous velocity commands.

\section{Image-Goal Navigation via Reinforcement Learning}
\subsection{Task Definition} 
Given a goal RGB image in an unknown environment, the agent must search the area for the target using panoramic observations. At each step $t$, the agent samples an action $a_t$ conditioned on visual observations $s_t$, and gets a reward $r_t$. The policy adopts a continuous action space $a_t \in \mathcal{A}_c = \{v_{\text{lin}}, v_{\text{ang}}\} $ consisting of angular and forward linear velocities in the body frame. The task is considered successful if the distance to the goal viewpoint is less than $1~m$.

\subsection{Navigation Model} \label{sec:nav_model}
\begin{figure}[!t]
    \centering
    \includegraphics[width=\linewidth]{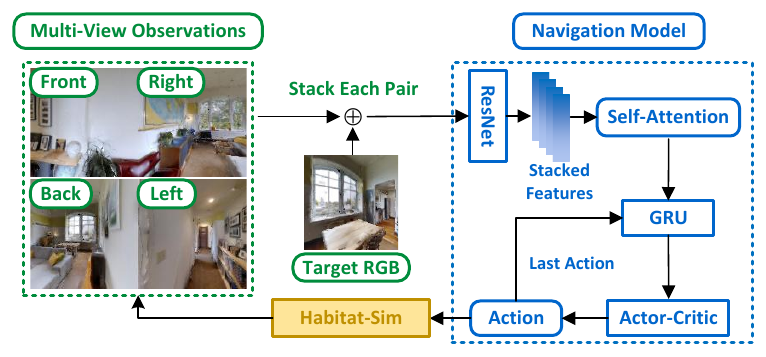}
    \caption{Overview of the navigation model. The RGB image from each direction is stacked with the target image and encoded by a shared ResNet. The resulting per-view features are adaptively fused by a self-attention module and passed to a GRU together with the previous action. The actor-critic head then predicts continuous actions.}
    \label{fig:model}
\end{figure}

As illustrated in Fig.~\ref{fig:model}, the navigation model takes four RGB views $O_c=\{O_c^{k}\}_{k=1}^{4}$ (front, right, back, left) together with the goal image $O_g$ as raw inputs. Each view is separately stacked with $O_g$ to obtain four pairs, which are encoded by a ResNet encoder $f(\cdot)$ to produce a per-view feature map
\begin{equation}
    s_t^{k} = O_c^{k} \oplus O_g, \quad F_t^{k} = f(s_t^{k}), \quad k=1,\dots,4.
\end{equation}
Rather than directly concatenating the four feature maps, we introduce a self-attention module to fuse latent features across views adaptively. This enables the policy to emphasize views that are most relevant to the goal image. Since each view corresponds to a fixed yaw angle, the learned attention also provides implicit directional guidance for action selection. Specifically, each feature map is first global-average-pooled (GAP) into a compact token $e_t^{(k)} \in \mathbb{R}^{C}$. The four tokens are then stacked and layer-normalized (LN):
\begin{equation}
    e_t^{k} = \mathrm{GAP}\big(F_t^{k}\big), \quad
    E_t = \mathrm{LN}\big([e_t^{1}, \dots, e_t^{4}]^{\top}\big)
    \in \mathbb{R}^{4 \times C}.
\end{equation}
Multi-head self-attention with $H$ heads is then applied to capture cross-view interactions:
\begin{equation}
\begin{aligned}
    \mathrm{head}_i &=
    \mathrm{softmax}\!\left(
    \frac{(E_t W_i^{Q})(E_t W_i^{K})^{\top}}{\sqrt{C/H}}
    \right)(E_t W_i^{V}), \\
    \tilde{E}_t &=
    [\mathrm{head}_1, \dots, \mathrm{head}_H] W^{O},
\end{aligned}
\end{equation}
where $W_i^{Q}, W_i^{K}, W_i^{V} \in \mathbb{R}^{C \times (C/H)}$ and
$W^{O} \in \mathbb{R}^{C \times C}$ are learnable projection matrices.
Let $\tilde{e}_t^{(k)}$ denote the attended token of the $k$-th view, corresponding to the $k$-th row of $\tilde{E}_t$. A shared linear gating layer maps each $\tilde{e}_t^{(k)}$ to a scalar score. A softmax over the four scores then produces adaptive view weights:
\begin{equation}\label{eq:view_weights}
    s_t^{(k)} = \tilde{e}_t^{(k)} W_g + b,
    w_t^{(k)} =
    4 \cdot
    \frac{\exp(s_t^{(k)})}
    {\sum_{j=1}^{4}\exp(s_t^{(j)})}, k=1,\dots,4,
\end{equation}
where $W_g \in \mathbb{R}^{C \times 1}$ and $b \in \mathbb{R}$ are
learnable parameters. The weighted feature maps are then concatenated as
\begin{equation}
    h_t =
    [w_t^{(1)}F_t^{(1)},\, \dots,\, w_t^{(4)}F_t^{(4)}].
\end{equation}

A Gated Recurrent Unit (GRU) network $g(\cdot)$ integrates the fused visual representation $h_t$ with the previous action $a_{t-1}$ to obtain the hidden state
$z_t$. The actor-critic policy then predicts the action distribution
and state value:
\begin{equation}
    z_t = g(h_t, a_{t-1}), \
    a_t \sim \pi_{\theta}(\cdot \mid z_t), \
    v_t = V_{\phi}(z_t).
\end{equation}

The policy is optimized by Proximal Policy Optimization (PPO)~\cite{ppo}. The clipped policy objective and value loss are given by
\begin{equation}
\begin{aligned}
    L_{a}(\theta)
    &=
    \mathbb{E}_t
    \left[
    \min\!\left(
    r_t(\theta)\hat{A}_t,\,
    \mathrm{clip}\!\left(
    r_t(\theta),1-\epsilon,1+\epsilon
    \right)\hat{A}_t
    \right)
    \right], \\
    L_{v}(\phi)
    &=
    \mathbb{E}_t
    \left[
    \frac{1}{2}
    \left(
    \hat{R}_t - V_{\phi}(z_t)
    \right)^2
    \right],
\end{aligned}
\end{equation}
where $r_t(\theta)=
    \frac{\pi_{\theta}(a_t \mid z_t)}
    {\pi_{\theta_{\mathrm{old}}}(a_t \mid z_t)}$,
$\hat{A}_t$ is computed using generalized advantage estimation, and
$\hat{R}_t$ denotes the return target for the value function.

\begin{figure*}[!t]
    \centering
    \includegraphics[width=\linewidth]{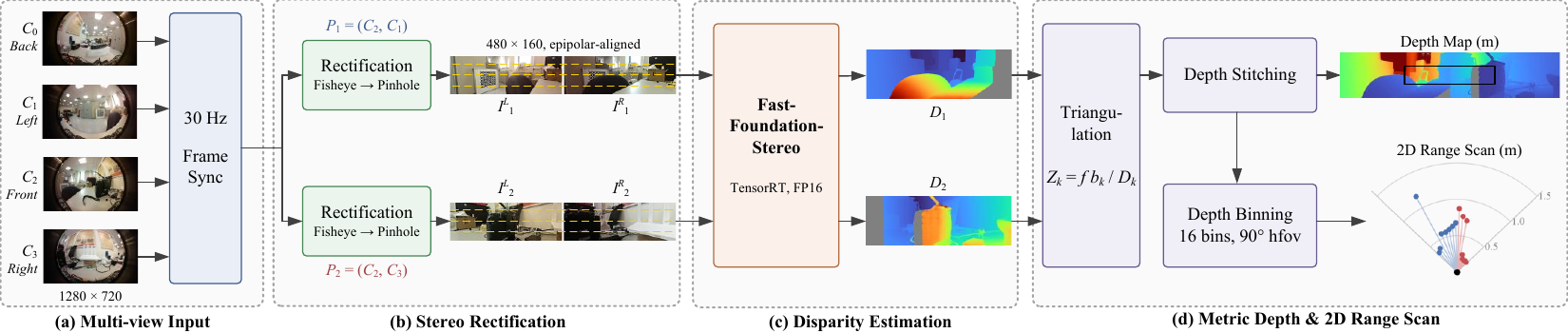}
    \caption{Stereo depth estimation pipeline. (a) Fisheye images are frame-synchronized and captured at 30 Hz. (b) The front camera forms two stereo pairs, $\mathcal{P}_1=(C_2, C_1)$ and $\mathcal{P}_2=(C_2, C_3)$, with the left and right cameras, respectively. Each pair is rectified into aligned pinhole images. (c) Fast-FoundationStereo produces disparity maps $D_1$ and $D_2$ for both pairs. (d) The disparity maps are converted to depth and stitched together, then reduced over the central $90^\circ$ FoV to a 16-bin range scan for collision prediction.}
    \label{fig:depth}
\end{figure*}

\subsection{Reward Design and Evaluation Metrics}
The reward function encourages the policy to approach the target viewpoint and align the front view with the goal view. Following standard ImageNav benchmark settings, the goal is specified by a single-view RGB image, while the policy receives observations from four cameras. The reward at time $t$ is defined as
\begin{equation}
    r_t = (d_{t-1}-d_t) + \Phi_t+ r_{\mathrm{suc}}m_t
    + r_{\mathrm{slack}},
\end{equation}
where $d_t$ denotes the distance to the goal, and the progress term $d_{t-1}-d_t$ encourages movement toward the goal. We set the slack penalty to $r_{\mathrm{slack}}=-0.01$ and the success reward to $r_{\mathrm{suc}}=+5.0$. The orientation error 
\begin{equation}
\begin{aligned}
     \Phi_t &= e_{t-1} - e_t, \quad
    e_t =
    \begin{cases}
        \alpha_t, & d_t \leq d_s, \\
        \pi, & d_t > d_s.
    \end{cases}
\end{aligned}
\end{equation}
encourages the agent to match the goal viewpoint when near the target. Here, $\alpha_t$ is the angular deviation between the current front view and the goal view. We set the distance and angular thresholds to $d_s=1~\mathrm{m}$ and $\alpha_s=25^\circ$, respectively.
The goal-matching coefficient $m_t$ is defined as 
\begin{equation}
\begin{aligned}
    m_t &= \mathbb{I}[d_t \leq d_s]
    \left(0.5+0.5\mathbb{I}[\alpha_t \leq \alpha_s]\right), \\
    \mathbb{I}[x] &=
    \begin{cases}
        1, & \text{if } x \text{ is true}, \\
        0, & \text{otherwise}.
    \end{cases}
\end{aligned}
\end{equation}
Satisfying the distance criterion yields a bonus of $2.5$, which increases to $5.0$ when the angular criterion is also satisfied.

We evaluate navigation performance using Success Rate (SR) and Success weighted by Path Length (SPL):
\begin{equation}
    \mathrm{SR} = \frac{1}{N}\sum_{i=1}^{N}S_i,
    \quad
    \mathrm{SPL} = \frac{1}{N}\sum_{i=1}^{N}
    S_i\frac{L_i}{\max(L_i,p_i)},
\end{equation}
where $N$ is the number of evaluation episodes, $S_i\in\{0,1\}$ indicates whether episode $i$ is successful, $p_i$ denotes the executed path length, and $L_i$ is the shortest navigable path length from the start position to the goal. SR measures overall task completion, while SPL accounts for path efficiency.

\section{Depth Estimation for Collision Avoidance} \label{chapter:4}
Habitat benchmarks encourage goal reaching but do not penalize collisions, so the trained navigation policy is not safety-aware. To ensure safe deployment, we apply a safety layer~\cite{usl} to raw navigation actions. The depth-based action correction mechanism intervenes only when a potential collision is detected. However, our drone platform (Fig.~\ref{fig:title}) is equipped only with quad-fisheye RGB cameras and does not provide direct depth measurements. To address this limitation, we develop a real-time stereo depth estimation pipeline for safety-aware action correction.

\begin{figure}[!t]
    \centering
    \includegraphics[width=\linewidth]{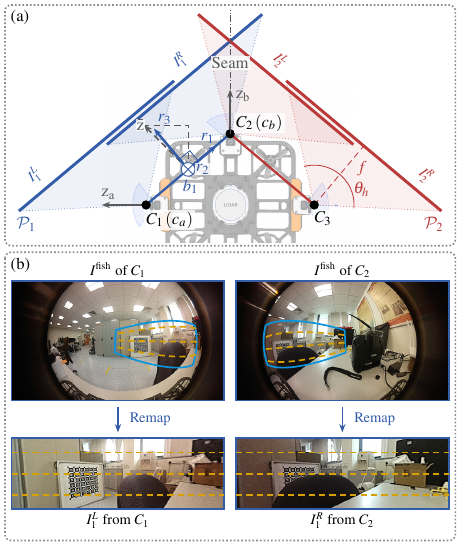}
    \caption{Stereo rectification of the quad-fisheye views. (a) Rectified frame construction for the two stereo pairs. For $\mathcal{P}_1$, $C_1$ and $C_2$ supply the left and right rectified views, respectively. (b) Real fisheye observations and their rectified views for $\mathcal{P}_1$. The dashed curves in the fisheye images are mapped to aligned image rows.}
    \label{fig:rectify}
\end{figure}

\subsection{Learning-Based Stereo Depth Estimation} \label{sec:depth}
As shown in Fig.~\ref{fig:depth}, we estimate stereo depth from quad-fisheye images to support collision avoidance. Since the navigation policy outputs forward linear and angular velocities, obstacle perception focuses on the frontal region. The front camera $C_2$ is paired with the adjacent left and right cameras $C_1$ and $C_3$, yielding $\mathcal{P}_1=(C_2,C_1)$ and $\mathcal{P}_2=(C_2,C_3)$. The overlapping FoVs provide stereo observations for recovering metric depth.

For each stereo pair, two fisheye images are rectified into a shared virtual pinhole frame, as illustrated in Fig.~\ref{fig:rectify}. For $\mathcal{P}_1$, let $c_a,c_b$ and $R_a,R_b$ denote the centers and calibrated world-to-camera rotations of $C_1$ and $C_2$, respectively.
We first align the horizontal axis of the rectified frame with the stereo baseline:
\begin{equation}
    r_1 =
    \frac{c_b-c_a}
    {\|c_b-c_a\|_2}.
\end{equation}
The optical-axis directions of the two cameras in the world frame are given by
\begin{equation}
    z_a = R_a^\top [0,0,1]^\top,
    z_b = R_b^\top [0,0,1]^\top,
    \bar{z}=z_a+z_b.
\end{equation}
The forward axis $r_3$ is obtained by projecting $\bar{z}$ onto the plane perpendicular to $r_1$:
\begin{equation}
    r_3 =
    \frac{
    \bar{z}
    -[\bar{z}^\top r_1]r_1
    }{
    \left\|
    \bar{z}
    -[\bar{z}^\top r_1]r_1
    \right\|_2
    },
    \quad
    r_2 = r_3 \times r_1.
\end{equation}
The resulting rectification rotation is $R_{\mathrm{rect}}=[r_1,r_2,r_3]^\top.$
Both stereo pairs share the same rectified intrinsic matrix
\begin{equation}
    K_{\mathrm{rect}}=
    \begin{bmatrix}
        f & 0 & c_x\\
        0 & f & c_y\\
        0 & 0 & 1
    \end{bmatrix},
\end{equation}
where
\begin{equation}
    f=\frac{W/2}{\tan(\theta_h/2)},
    \quad
    c_x=\frac{W-1}{2},
    \quad
    c_y=\frac{H-1}{2}.
\end{equation}
Here, $W \times H$ denotes the rectified image resolution and $\theta_h$ is the horizontal FoV of the virtual pinhole camera. We use $W=480$, $H=160$, and $\theta_h=110^\circ$ for fast onboard inference.
For each pixel $p=[u,v,1]^\top$ in the rectified image, the corresponding pixel $q_c$ in the fisheye image of camera $c\in\{a,b\}$ is given by
\begin{equation}
q_c =
\pi_c\left(
R_c R_{\mathrm{rect}}^\top
K_{\mathrm{rect}}^{-1}p
\right), \
I_c^{\mathrm{rect}}(p)
= I_c^{\mathrm{fish}}(q_c).
\end{equation}
Here, $\pi_c(\cdot)$ is the calibrated fisheye projection model. The remapping tables are precomputed and applied to each incoming frame to obtain the epipolar-aligned pair, e.g., $(I^L_1, I^R_1)=(I^{\mathrm{rect}}_a, I^{\mathrm{rect}}_b)$ in Fig.~\ref{fig:rectify}.

We use Fast-FoundationStereo~\cite{ffs} to estimate disparity maps $D_1$ and $D_2$ from the rectified stereo pairs $(I^L_k, I^R_k)$. Each disparity map is converted to metric depth by stereo triangulation:
\begin{equation}
Z_k=\frac{f b_k}{D_k}, \quad
b_k=\left\|c_b^{(k)}-c_a^{(k)}\right\|_2, \quad
k\in\{1,2\},
\end{equation}
where $f$ is the focal length, $b_k$ is the baseline of stereo pair $\mathcal{P}_k$ and $Z_k$ is depth along its rectified optical axis. The estimated stereo depth maps are stitched along a geometric seam to form a unified depth map of the frontal region.

\subsection{Obstacle Avoidance via Action Safety Layer}
\begin{figure}[!t]
    \centering
    \includegraphics[width=\linewidth]{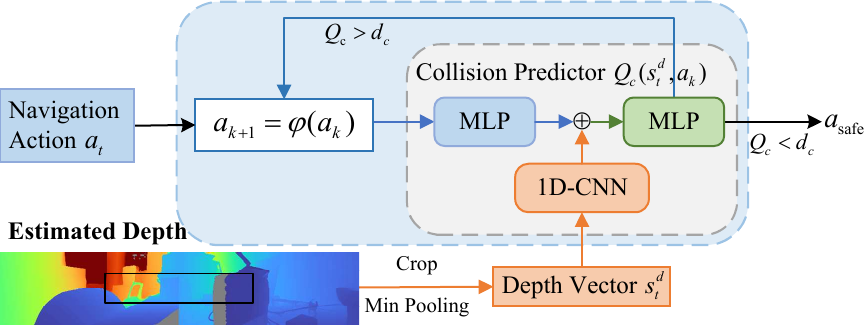}
    \caption{Action safety layer based on estimated depth for collision avoidance.}
    \label{fig:correct}
\end{figure}

Fig.~\ref{fig:correct} illustrates the action safety layer. Given the current depth vector $s_t^d\in\mathbb{R}^{16}$ and normalized policy action $a_t$, the collision predictor $Q_c(s_t^d,a_t) \in [0,1]$ predicts the collision probability.
To construct $s_t^d$, we select the central $90^\circ$ horizontal FoV of the stitched depth map and a central image band spanning 40\% of its height, as illustrated in Fig.~\ref{fig:depth}(d). We divide the selected depth region into 16 horizontal bins and retain the minimum valid range in each bin.  Finally, the ranges are clipped at 3~m and normalized to $[0,1]$. The collision predictor $Q_c$ is inherited from \cite{yan2025sign}. 
If $Q_c(s_t^d,a_t)\leq d_c$, the action is passed through unchanged. Otherwise, the safety layer iteratively adjusts the action via gradient descent to reduce the predicted risk, starting from $a_0=a_t$:
\begin{equation}\label{USL}
    a_{k+1}=\varphi(a_k)
    =a_k-\nabla_{a_k}\left[Q_c(s_t^d,a_k)-d_c\right]^+,
\end{equation}
where $[x]^+=\max(x,0)$ and $k$ indexes the correction iterations. Correction stops when $Q_c(s_t^d,a_k)\leq d_c$ or the iteration limit is reached. The threshold $d_c$ controls the tradeoff between safety and navigation efficiency, with lower values yielding more conservative action correction.

\section{Simulation Experiments}
\subsection{Simulation Settings}
\subsubsection{Datasets} 
Our ImageNav model is trained on Gibson dataset \cite{gibson}, including 72 training scenes and 14 testing scenes. To further validate the robustness of our algorithm, the trained model is also evaluated on cross-domain datasets MP3D \cite{mp3d} and HM3D \cite{hm3d}. Scenes from different datasets have varying backgrounds and room layouts. 

\subsubsection{Agent Configurations}
To approximate the quad-fisheye rig in simulation, we equip the agent with four pinhole RGB cameras mounted at $90^\circ$ intervals around the agent (front, right, back, left), jointly covering a full $360^\circ$ panorama. Each RGB image retains the same $90^\circ$ FoV and $128 \times 128$ resolution used by prior works. The continuous action space $\mathcal{A}_c$ consists of forward linear velocity $v_{\text{lin}} \in [0,0.25]~m/s$ and angular velocity $v_{\text{ang}} \in [-15, 15]~^\circ/s$. To implement the \textit{stop} action in a continuous setting, the episode will be terminated early when $v_{\text{lin}}<0.025~m/s \ \& \ |v_{\text{ang}}|<1.5~^\circ/s$.

\subsubsection{Training Details}
To support fast inference and efficient onboard deployment, Omni-IGN uses ResNet9 as a lightweight backbone. The visual encoder is initialized with pretrained weights from \cite{yan2025sign} as a warm start. Then we train the full model with PPO for 50M steps on the Gibson benchmark using an NVIDIA H200 GPU. Training uses $B=8$ parallel environments, each collecting 512 steps per rollout, yielding a total rollout batch size of 4096.

\begin{figure*}[!t]
    \centering
    \includegraphics[width=\linewidth]{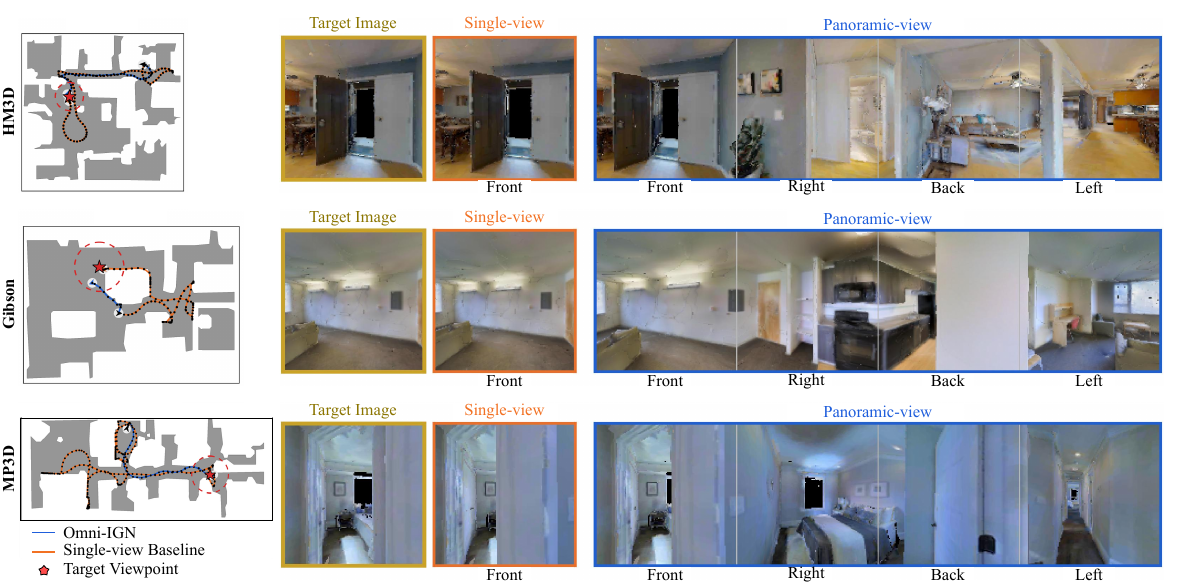}
    \caption{Comparison experiments between \textcolor{blue}{Omni-IGN} and  \textcolor{orange}{single-view baseline} on the same episodes from HM3D, Gibson, and MP3D. Left: navigation trajectories on the top-down map. Right: the goal image, the single-view and panoramic-view observations at the endpoint.}
    \label{fig:4rgb}
\end{figure*}

\subsection{Performance Analysis on Benchmarks}
\subsubsection{Evaluation Results}
On simulation benchmarks, the navigation policy is evaluated using SR and SPL without considering safety. Table~\ref{table:gibson} compares Omni-IGN with other SOTA ImageNav methods on Gibson. Compared with SIGN, Omni-IGN improves SR by \textbf{+1.5\%} and SPL by \textbf{+10.9\%}. It also outperforms 4-RGB FGPrompt by \textbf{+3.3\%} in SR and \textbf{+10.1\%} in SPL under continuous control. Omni-IGN achieves performance comparable to the discrete-action SOTA while enabling smoother and continuous flight control without requiring a map.

\begin{table}[!t]
    \caption{Comparison with SOTA methods on Gibson.\label{table:gibson}}
    \centering
    \setlength{\tabcolsep}{3pt}
    \begin{tabular}{ccccccc}
    \hline Method & Backbone & Sensor(s) & Action Space & SPL($\uparrow$) & SR($\uparrow$) \\
    \hline
    ZER \cite{zer} & ResNet9  & RGB & Discrete & 21.6\% & 29.2\% \\
    OVRL-V2 \cite{ovrlv2} & ViT-Base  & RGB & Discrete &58.7\% & 82.0\% \\
    FGPrompt \cite{fgp} & ResNet9  & RGB & Discrete & 66.5\% & 90.4\% \\
    Mem-Aug \cite{memaug} & ResNet18  & 4 RGB & Discrete & 56.0\% & 69.0\% \\
    VGM \cite{vgm} & ResNet18  & 4 RGB & Discrete & 64.0\% & 76.0\% \\
    TSGM \cite{tsgm} & ResNet18  & 4 RGB & Discrete & 67.2\% & 81.1\% \\
    \hline 
    FGPrompt \cite{fgp} & ResNet9  & RGB & Continuous & 35.0\% & 79.9\% \\
    SIGN \cite{yan2025sign} & ResNet9 & RGB & Continuous &  53.3\% & 86.3\% \\
    FGPrompt \cite{fgp} & ResNet9  & 4 RGB & Continuous & 54.1\% & 84.5\% \\
    Omni-IGN w/o SA & ResNet9 & 4 RGB & Continuous & 63.1\% & 84.5\% \\
    \rowcolor{gray!20}
    Omni-IGN & ResNet9 & 4 RGB & Continuous & 64.2\% & 87.8\% \\
    \hline
    \end{tabular}
\end{table}

\begin{table}[!t]
    \caption{Cross-domain evaluation on MP3D and HM3D.\label{table:cross}}
    \centering
    \setlength{\tabcolsep}{3pt}
    \begin{tabular}{cccccc}
    \hline \multirow{2}{*}{ Methods } & \multirow{2}{*}{ Action Space } & \multicolumn{2}{c}{ MP3D } & \multicolumn{2}{c}{ HM3D } \\
    \cline { 3 - 6 } & & SPL($\uparrow$) & SR($\uparrow$) & SPL($\uparrow$) & SR($\uparrow$) \\
    \hline 
    ZER \cite{zer} & Discrete& $10.8\% $ & $14.6\% $ & $6.3\% $ & $9.6\% $ \\
    TSGM \cite{tsgm} & Discrete& $47.3\%$ & $58.0\% $ & $46.9\% $ & $64.1\%$ \\
    FGPrompt (1-RGB) & Discrete& $50.4\% $ & $77.6\% $ & $49.6\% $ & $76.1\% $ \\
    \hline
    FGPrompt (1-RGB) & Continuous& $22.9\% $ & $59.4\% $ & $20.1\% $ & $56.5\% $ \\
    SIGN \cite{yan2025sign} & Continuous& $35.0\%$ & $65.4\% $ & $30.9\% $ & $64.5\%$ \\
    FGPrompt (4-RGB) & Continuous& $42.2\%$ & $70.1\% $ & $35.0\% $ & $66.8\%$ \\
    Omni-IGN w/o SA & Continuous& $46.2\%$ & $68.0\% $ & $38.4\% $ & $64.0\%$\\
    \rowcolor{gray!20}
    Omni-IGN & Continuous& $45.1\%$ & $70.6\% $ & $38.8\% $ & $69.2\%$\\
    \hline
    \end{tabular}
\end{table}

To assess cross-domain generalization, we directly evaluate Gibson-trained policies on MP3D and HM3D without fine-tuning. Results are averaged across easy, medium, and hard settings, each containing 1000 episodes. As shown in Table~\ref{table:cross}, 4-RGB FGPrompt achieves the best performance among continuous baselines. On MP3D, Omni-IGN improves SR by +0.5\% and +2.9\% in SPL. On HM3D, Omni-IGN surpasses 4-RGB FGPrompt by +2.4\% in SR and +3.8\% in SPL. Overall, our method achieves the highest SR under continuous control across all three benchmarks.

\begin{figure*}[!t]
    \centering
    \includegraphics[width=0.92\linewidth]{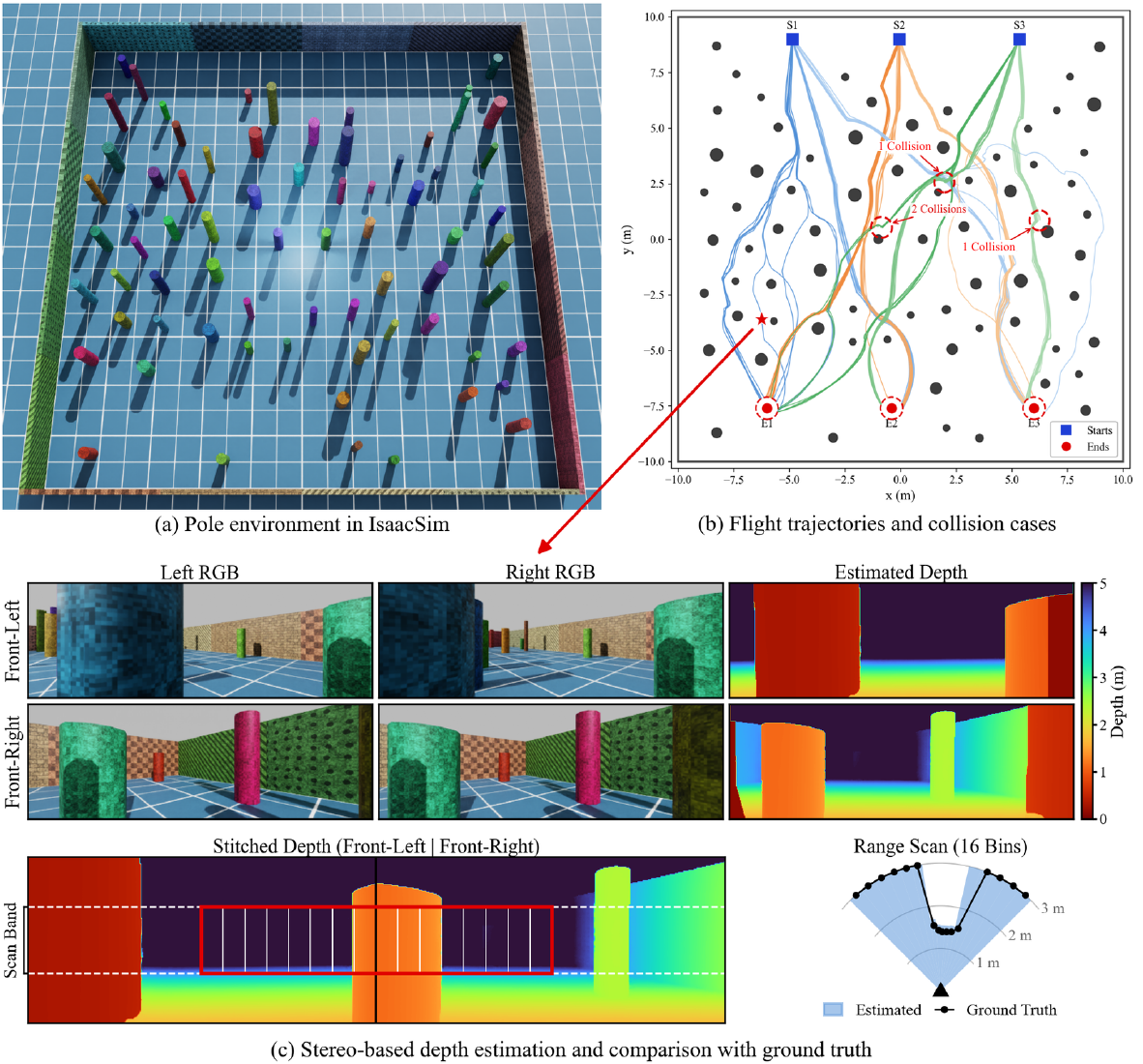}
    \caption{Reliability test in the IsaacSim pole environment. (a) The layout of the pole environment. (b) Flight trajectories for each start-end pair, with collision cases marked in red. (c) Stereo-based depth estimation at the location marked by the red star in (b). Top: left and right RGB views of the front-left and front-right stereo pairs and their estimated depth. Bottom: the two depth maps are stitched and cropped to form the range scan.}
    \label{fig:pole_room}
\end{figure*}

\subsubsection{Benefits of Omnidirectional Perception}
Compared with the single-view baseline using the same continuous action space, Omni-IGN achieves a substantially larger improvement in SPL than in SR on benchmarks. This suggests that broader visual coverage reduces the need for exploration, leading to more efficient navigation paths, while also improving SR by reducing the likelihood of missing the goal image.

Fig.~\ref{fig:4rgb} compares the performance in the same episodes across all three benchmarks. The single-view baseline often requires repeated exploration, resulting in longer trajectories with additional loops and backtracking. In contrast, our method takes more direct routes toward the target viewpoint. These cases illustrate why omnidirectional perception can improve SPL.

\subsubsection{Ablation of Self-Attention}
To assess the contribution of self-attention (SA), we compare Omni-IGN with a variant without SA in Tables~\ref{table:gibson} and~\ref{table:cross}. On Gibson, adding SA improves SR and SPL by +3.3\% and +1.1\%, respectively. Without SA, the panoramic policy achieves an SPL gain of +9.8\% over SIGN, but an SR change of -1.8\%. It means that although a broader visual horizon can reduce unnecessary exploration, simply combining views without adaptive weighting may introduce additional distractions or interfere with precise goal reaching. 
In cross-domain evaluation, adding SA improves SR by +2.6\% on MP3D and +5.2\% on HM3D, while maintaining comparable SPL.

\subsection{Reliability Analysis of Depth Estimation and Safety Layer}
Obstacle avoidance is built on the collision predictor $Q_c$ and the safety layer. To evaluate their robustness and integration with the PX4 autopilot, we conduct repeated flight experiments in IsaacSim, which approximates real-world flight dynamics and provides ground-truth depth for quantitative analysis. As shown in Fig.~\ref{fig:pole_room}, we define a set of start and end positions. For each start-end pair, 10 repeated point-to-point flight tests are conducted, where a default PX4 controller tracks reference velocities $v_{\text{ang}}$ and $v_{\text{lin}}$. The default $v_{\text{lin}}$ is set to $0.25~m/s$, which matches the maximal speed during ImageNav training. The target $v_{\text{ang}}$ is computed based on the yaw deviation between the current orientation and the direction toward the goal. A trial is considered successful if the drone reaches within $0.5~m$ of the goal without any collision.

\begin{table}[!t]
\centering
\caption{Reliability test of collision avoidance and depth estimation over 90 flights in IsaacSim.}
\label{tab:pole_room}
\setlength{\tabcolsep}{3pt}
\footnotesize
\begin{tabular}{@{}lcccccc@{}}
\toprule
 & \multicolumn{3}{c}{Navigation} & \multicolumn{2}{c}{Depth Error (m)} \\
\cmidrule(lr){2-4} \cmidrule(l){5-6}
Pair & SR($\uparrow$) & SPL($\uparrow$) & Correction Rate(\%) & Mean($\downarrow$) & Max($\downarrow$) \\
\midrule
S1-E1 & 10/10 & 0.93  & 30.4 & 0.17 & 1.13 \\
S1-E2 & 10/10 & 0.96  & 14.0 & 0.10 & 0.80 \\
S1-E3 & 10/10 & 0.91  & 28.8 & 0.17 & 1.09 \\
S2-E1 & 10/10 & 0.99  & 18.1 & 0.13 & 1.04 \\
S2-E2 & 10/10 & 0.94  & 23.5 & 0.13 & 1.00 \\
S2-E3 & 10/10 & 0.96  & 24.5 & 0.16 & 1.00 \\
S3-E1 & 8/10 & 0.77  & 24.7 & 0.14 & 0.92 \\
S3-E2 & 9/10 & 0.83  & 22.1 & 0.15 & 0.92 \\
S3-E3 & 9/10 & 0.88  & 23.2 & 0.14 & 1.03 \\
\midrule
Average & 0.956 & 0.91 & 23.3 & 0.14 & 0.99 \\
\bottomrule
\end{tabular}
\end{table}

In our experiments, the maximum depth is clipped at $3~\mathrm{m}$ and normalized to $[0,1]$. We convert the estimated depth into a range scan $s_t^d \in \mathbb{R}^{16}$ as input to $Q_c$. In Eq.~\eqref{USL}, we set the collision probability threshold to $d_c=0.65$ and iteratively correct actions until $Q_c < d_c$. Instead of using gradient descent, we adopt a more practical and empirical action-correction scheme, since gradient-based optimization can produce oscillatory action updates and compromise planning consistency under noisy risk estimates,
\begin{equation}
    \varphi(a_{k}) = a_k-\Delta a = (v_{\text{lin}}-\Delta_{\text{lin}}, v_{\text{ang}}-D \cdot \Delta_{\text{ang}}),
\end{equation}
where $D \in \{-1,1\}$ denotes the direction with greater depth, either left or right. For a normalized $v_{\text{lin}},v_{\text{ang}} \in [-1,1]$, $\Delta_{\text{lin}}=0.3, \Delta_{\text{ang}}=0.3$ mean deceleration and turning toward the obstacle-free side. To maintain real-time inference, we allow at most five correction iterations; otherwise, the linear speed will be set to 0. Such a design helps the drone rotate in place at $v_{\text{ang}}$ to escape the local dead end.

\begin{figure*}[!t]
    \centering
    \noindent\makebox[\linewidth][l]{(I) Navigation to a goal image featuring a door and a red cone.}
    \par\smallskip
    \includegraphics[width=\linewidth]{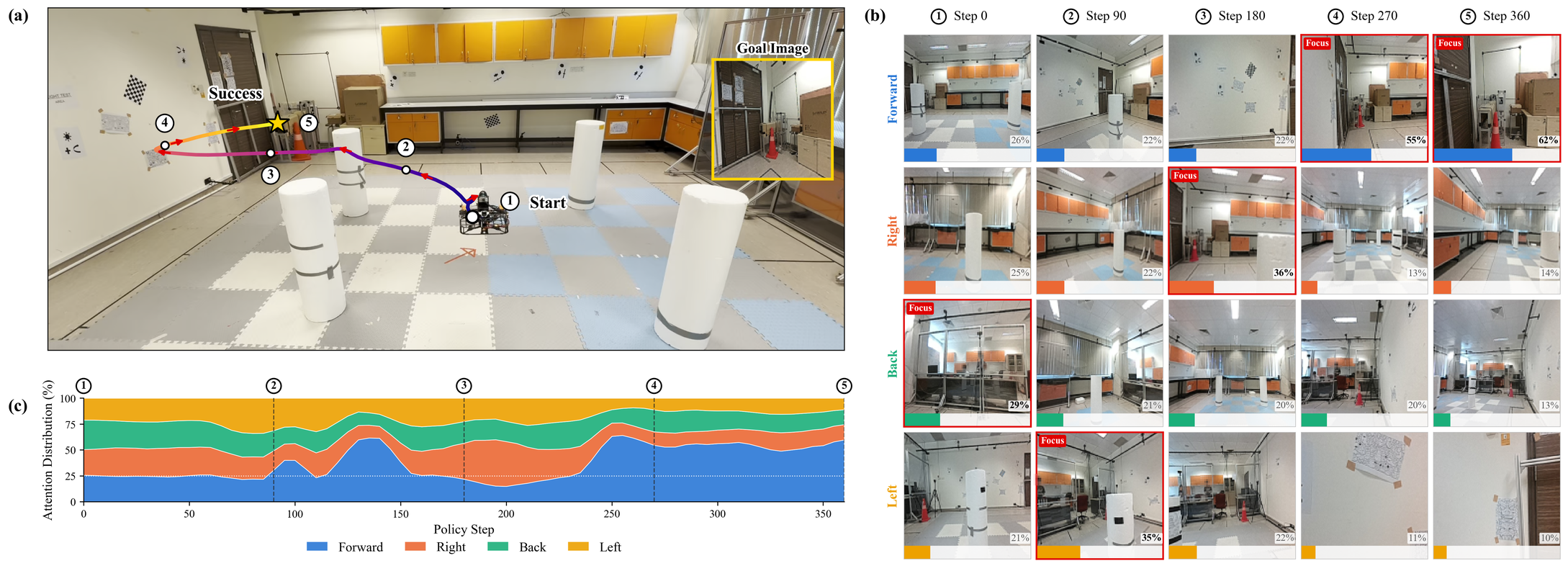}
    \par\medskip
    \noindent\makebox[\linewidth][l]{(II) Navigation to a goal image featuring blue blocks.}
    \par\smallskip
    \includegraphics[width=\linewidth]{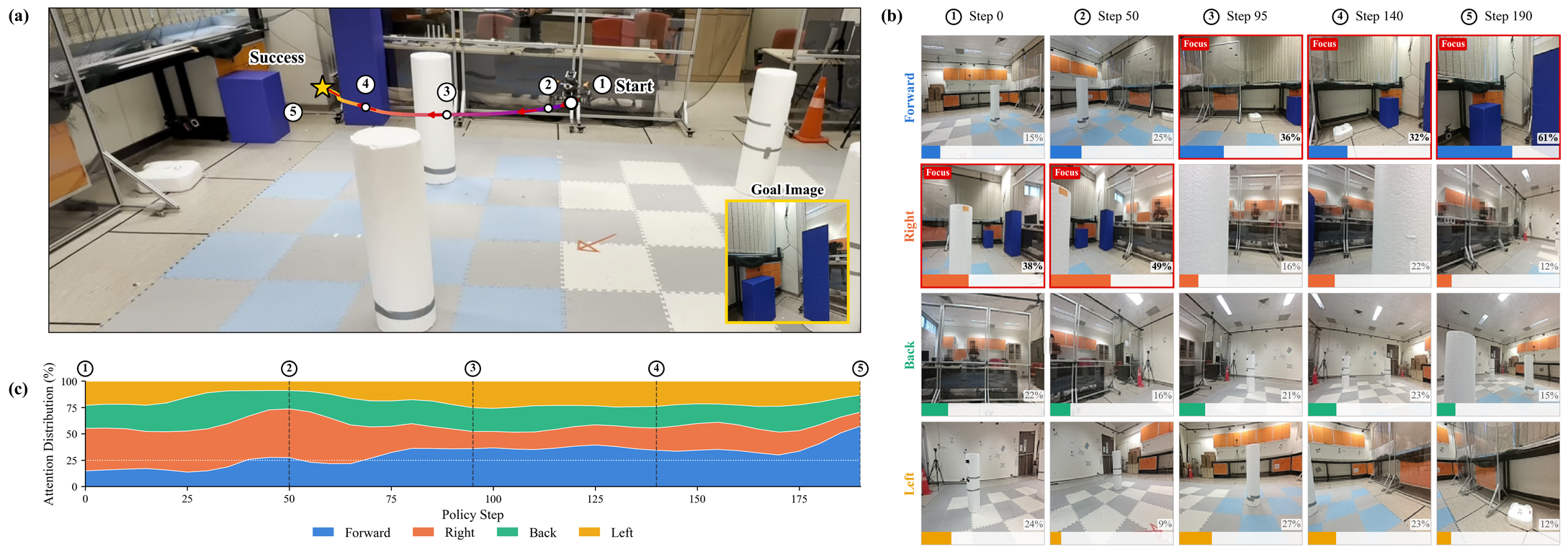}
    \par
    \caption{Evolution of view-attention weights during two real-world ImageNav flights in the Lab. Panels (I-a) and (II-a) show the trajectories and goal images. (I-b) and (II-b) show the four camera views and normalized attention scores at five sampled steps. (I-c) and (II-c) show the attention distributions over time. Red boxes highlight the view with the highest attention score.}
    \label{fig:att_office}
\end{figure*}

\begin{figure*}[!t]
    \centering
    \includegraphics[width=\linewidth]{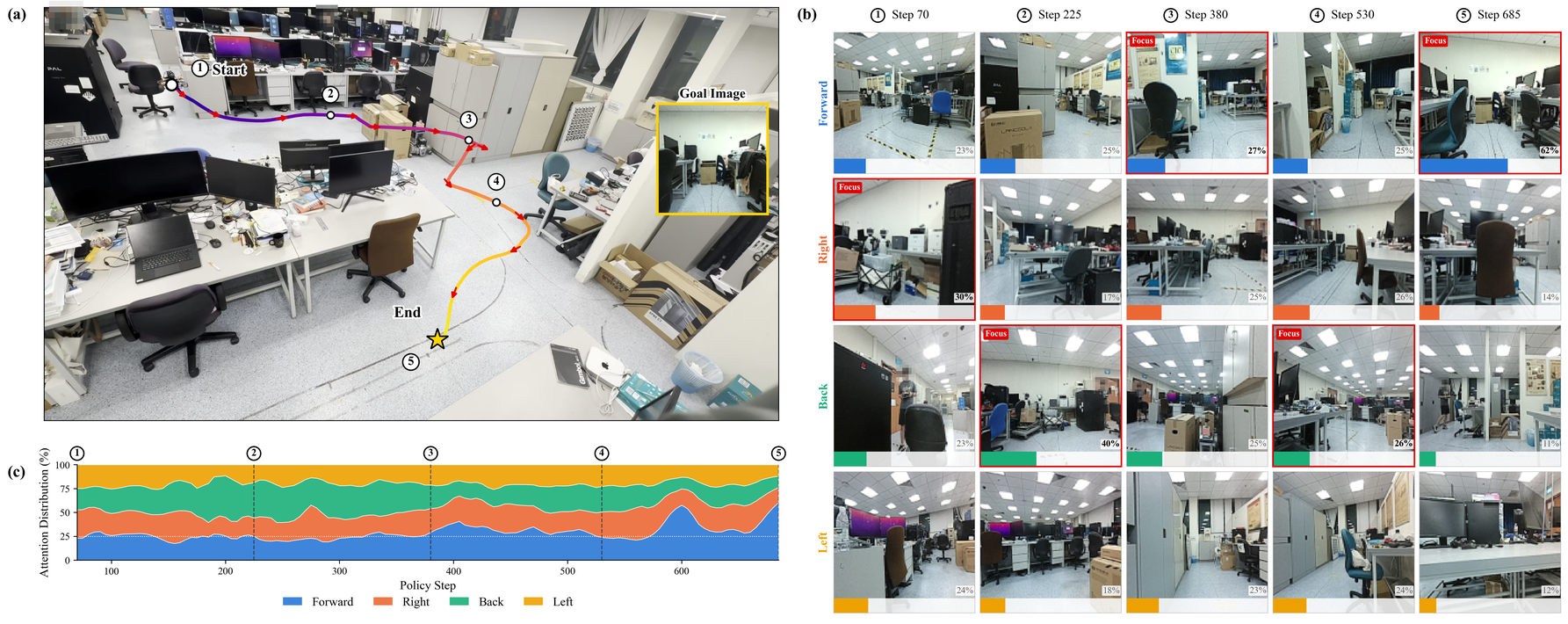}
    \caption{Evolution of view attention during a real-world ImageNav flight in a cluttered office. Attention is broadly distributed during exploration and becomes increasingly focused on the forward view during the final approach.}
    \label{fig:att_lab}
\end{figure*}

Fig.~\ref{fig:pole_room}(b) and Table~\ref{tab:pole_room} summarize the experimental results. As shown in Fig.~\ref{fig:pole_room}(b), the safety layer guides the drone to bypass obstacles. Across 90 flights, the drone reaches the goal in 86 cases, achieving a success rate of $95.6\%$ and an average SPL of $0.91$. The average action correction rate is $23.3\%$. All four failures are side collisions caused by the depth scan's limited FoV, especially when passing by the poles. 

Fig.~\ref{fig:pole_room}(c) compares the stereo-derived 16-bin range scan with the ray-cast ground truth. The estimated depth scan closely follows the ground truth, with a mean depth error of $0.14~\mathrm{m}$. The maximum depth error is $1.13~\mathrm{m}$. These results demonstrate that Fast-FoundationStereo~\cite{ffs} is capable of providing reliable depth estimation in real time. The estimation errors remain within an acceptable tolerance. With an accurate depth scan, the safety layer can effectively predict collision probability and correct unsafe actions, ensuring obstacle avoidance during autonomous navigation.

\section{Real-life Experiments}
\subsection{Design of Drone System}
\label{sec:system}

\subsubsection{Hardware Configuration}
As illustrated in Fig.~\ref{fig:title}, we conduct real-world experiments on a drone platform equipped with an NVIDIA Jetson Orin NX operating in MAXN mode. All perception, model inference, and control are executed fully onboard. An OAK-4P-New-B033501 frame-sync board with four fisheye cameras provides panoramic RGB observations at 30 Hz. The fisheye RGB streams provide input to both the ImageNav policy and stereo depth estimation.
Each fisheye lens has a FoV of $220^\circ$, and the four cameras are mounted at $90^\circ$ intervals, providing overlapping FoVs between adjacent cameras. This overlap allows the front camera to form stereo pairs with the left and right cameras for depth estimation, as described in Section~\ref{sec:depth}. For the ImageNav policy, each fisheye image is rectified into a $90^\circ$ pinhole view to match the camera configuration used during policy training, thereby reducing the sim-to-real gap in visual perception.

A Livox Mid-360s LiDAR is used only to provide state estimates for the low-level controller. Fast-lio2~\cite{fastlio2} runs at 10 Hz, and the PX4 autopilot fuses its output with IMU measurements to obtain reliable pose estimates. The velocity commands $u=[v_{\text{lin}},v_{\text{ang}}]$, defined in the body frame, are transformed to the world frame before being tracked by the PX4 controller.

\subsubsection{Acceleration of Onboard Computation}
To enable faster model inference, depth estimation and the ImageNav policy are accelerated using FP16 TensorRT engines. For stereo depth estimation, the $1280\times720$ fisheye images are rectified into $480\times160$ pinhole images with a horizontal FoV of $110^\circ$ and processed by Fast-FoundationStereo~\cite{ffs} using the \textit{23-36-37} checkpoint with \textit{valid\_iters}=8. The depth pipeline runs asynchronously in a separate worker process at approximately 3 Hz, without blocking policy execution. The ImageNav policy runs at approximately 10 Hz using rectified and resized $128\times128$ RGB images with a $90^\circ$ FoV.

\subsection{Verification of Attention Mechanism}
To examine how the self-attention module selects goal-relevant views, we visualize its per-view gating weights during three real-world flights, covering two stages: 1) navigation with visible goal cues and 2) exploration before the target scene becomes clearly visible. The weights in Eq.~\eqref{eq:view_weights} are normalized as $\alpha_t^{(k)}=w_t^{(k)}/4$, such that $\sum_{k=1}^{4}\alpha_t^{(k)}=1$. A uniform distribution corresponds to 25\% per view and leaves the feature maps equally weighted. A larger weight amplifies the corresponding feature before fusion, allowing the policy to emphasize the view with more goal-relevant cues.

Fig.~\ref{fig:att_office} presents two flights in the Lab where goal-relevant landmarks remain within the panoramic FoV throughout navigation. These examples show how attention tracks the target scene as the drone changes direction. In case I, the left view contains the traffic cone and cabinets visible in the goal image and receives the highest weight 35\% at step 90. As the drone turns, the landmark enters the right view, which receives the highest weight 36\% at step 180. During the final approach, the forward view captures the goal scene, and its weight increases to 55\% at step 270 and 62\% at step 360.
In case II, the blue blocks and surrounding structures in the goal image are initially visible in the right view. The distinctive blue color provides a salient cue and thus receives a high attention weight of 38\% at the first step. That's why the drone turns right instantly after taking off. Both flights demonstrate that the attention mechanism dynamically shifts toward goal-relevant views during navigation. During the final approach, the increasing attention weight on the forward view further proves that the model successfully recognizes the target.

Fig.~\ref{fig:att_lab} illustrates the attention behavior during early exploration in a cluttered office. When no single view provides a clearly distinctive goal cue, the attention weights remain broadly distributed across views, close to a uniform distribution. During exploration, attention is shared across the four views, with occasional preferences such as 40\% for the back view at step 225. Once the drone approaches the target viewpoint, the front weight rises to 62\% at step 685, while the remaining weights decrease to $11\%$--$14\%$. The shift from uniformly distributed attention to a dominant front view reflects the transition from exploration to goal-directed navigation. 

\subsection{Verification of Learning-Based Depth Estimation}
\begin{figure}[!t]
    \centering
    \includegraphics[width=\linewidth]{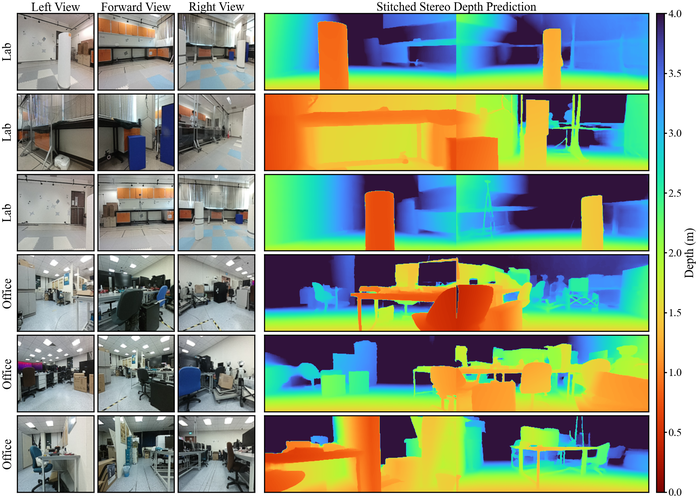}
    \caption{Real-world stereo depth estimation in the lab and office environments. The left, forward, and right RGB views are shown together with the corresponding stitched stereo depth predictions.}
    \label{fig:real_depth}
\end{figure}

Fig.~\ref{fig:real_depth} shows the RGB observations and stitched depth predictions in the lab and office environments. The learning-based stereo pipeline produces dense and spatially coherent depth estimates that clearly separate nearby obstacles from the background, including largely textureless cylindrical surfaces. Compared with raw depth measurements from conventional depth cameras, which often contain invalid regions, the predicted depth provides more complete spatial coverage. With accurate calibration of the fisheye cameras, learning-based depth estimates are sufficiently reliable for obstacle avoidance.

\section{Conclusions}
In this paper, we present Omni-IGN, a fully onboard aerial ImageNav system that leverages a quad-fisheye camera rig for both omnidirectional perception and safe navigation. Omni-IGN achieves SOTA performance among continuous-action baselines on Habitat benchmarks. Simulation and real-world experiments demonstrate that panoramic observations improve navigation efficiency and robustness. The whole system achieves zero-shot sim-to-real transfer without fine-tuning.

\bibliographystyle{IEEEtran}
\bibliography{IEEEabrv,reference}

\vfill

\end{document}